\documentclass[letterpaper, 10 pt, journal, twoside]{IEEEtran}
\IEEEoverridecommandlockouts                            
\usepackage{graphics} % for pdf, bitmapped graphics files
\graphicspath{ {./figures/}}
\usepackage{epsfig} % for postscript graphics files
\usepackage[figurename=Fig.]{caption}

\usepackage{times} % assumes new font selection scheme installed
\usepackage{amsmath} % assumes amsmath package installed
\usepackage{amssymb}  % assumes amsmath package installed
\usepackage{xcolor}
\usepackage{bm}
\usepackage{cancel}
\usepackage{color}
\usepackage{setspace}
\usepackage{wrapfig}
\usepackage{tikz}
\usetikzlibrary{shapes,arrows.meta,calc}
\usetikzlibrary{positioning}
\usepackage{caption}
\usepackage{subcaption}
\usepackage{booktabs}
\usepackage{multirow}
\usepackage{multicol}
\usepackage{soul}
\usepackage{cite}
\usepackage[normalem]{ulem}
\usepackage{xcolor}

\usepackage{enumitem}
\usepackage{tabularx}
\usepackage{courier}
\usepackage{algorithm}
\usepackage{algpseudocode} 
\usepackage{arydshln} % for \hdashline and dashed lines in tables
\usepackage{hyperref}

\usepackage{pifont}% http://ctan.org/pkg/pifont
\usepackage{url}

\newcommand{\todocite}[1]{\textcolor{red}{[TODO(cite)]}}

\allowdisplaybreaks

\newcommand{\re}[1]{{\color{black}#1}}

\title{\LARGE 
\textbf{
GRaD-Nav++: Vision-Language Model Enabled Visual Drone \underline{Nav}igation with \underline{G}aussian \underline{Ra}diance Fields and \underline{D}ifferentiable Dynamics
}
}

\author{Qianzhong Chen$^{*1}$, Naixiang Gao$^{*1}$, Suning Huang$^{2}$, JunEn Low$^{1}$, \\
Timothy Chen$^{2}$, Jiankai Sun$^{2}$, and Mac Schwager$^{2}$
\thanks{Manuscript received: July 4, 2025; Revised: October 7, 2025; Accepted: November 13, 2025.}
\thanks{This paper was recommended for publication by Editor Soon-Jo Chung upon evaluation of the Associate Editor and Reviewers' comments.}
\thanks{This work was supported in part by NSF grant 2342246 and ONR grant N00014-23-1-2354. 
Timothy Chen was partially supported on a NASA NSTGRO fellowship. Toyota Research Institute provided funds to support this work. 
(Corresponding author: Qianzhong Chen.)}
\thanks{$^{1}$ Authors are with the Department of Mechanical Engineering, Stanford University, Stanford, CA 94305, USA.
{\tt\footnotesize \{qchen23, nxiangao, jelow\}@stanford.edu}}
\thanks{$^{2}$ Authors are with the Department of Aeronautics and Astronautics, Stanford University, Stanford, CA 94305, USA.
{\tt\footnotesize \{suning, chengine, jksun, schwager\}@stanford.edu}}
\thanks{$^*$ Equal contributions.}
\thanks{Digital Object Identifier (DOI): see top of this page.}
}

\begin{document}

\maketitle
% for editing
% \thispagestyle{plain}
% \pagestyle{plain}
% %% for submission
% \thispagestyle{empty}
% \pagestyle{empty}

\begin{abstract}
Autonomous drones capable of interpreting and executing high-level language instructions in unstructured environments remain a long-standing goal. Yet existing approaches are constrained by their dependence on hand-crafted skills, extensive parameter tuning, or computationally intensive models unsuitable for onboard use. We introduce a lightweight Vision--Language--Action (VLA) framework that runs fully onboard and follows natural-language commands in real time. Our policy is trained in a photorealistic 3D Gaussian Splatting (3DGS) simulator via Differentiable Reinforcement Learning (DiffRL), enabling efficient learning of low-level control from visual and linguistic inputs. At its core is a Mixture-of-Experts (MoE) action head, which adaptively routes computation to improve generalization while mitigating forgetting. In multi-task generalization experiments, our method achieves a success rate of 83\% on trained tasks and 75\% on unseen tasks in simulation. When deployed on real hardware, it attains 67\% success on trained tasks and 50\% on unseen ones. In multi-environment adaptation experiments, our method achieves an average success rate of 81\% across diverse simulated environments and 67\% across varied real-world settings. These results establish a new benchmark for fully onboard Vision-Language-Action (VLA) flight and demonstrate that compact, efficient models can enable reliable, language-guided navigation without relying on external infrastructure. Project page: \url{https://qianzhong-chen.github.io/gradnavpp.github.io/}.
% \href{https://qianzhong-chen.github.io/gradnavpp.github.io/}{\textit{GRaD-Nav++}}.

\end{abstract}


% \begin{center}
%     SUPPLEMENTARY MATERIAL
% \end{center}
% Video: https://youtu.be/25Z7iAkZ5xw \\
% Code: https://github.com/HovakimyanResearch/L1-Mambo

\section{Introduction}
% Autonomous and intelligent drone
In recent years, autonomous drones have made remarkable progress in navigating complex environments, driven by advances in modular ``perception-planning-control" pipelines~\cite{gao2020teach, hanover2024autonomous}, imitation learning (IL)~\cite{low2024sous}, and reinforcement learning (RL)~\cite{chen2025grad,xu2025navrl,hu2025seeing}. These approaches have enabled drones to perform a broad range of tasks, ranging from basic waypoint following to sophisticated multi-agent coordination in dynamic environments. Despite these successes, existing methods are often restricted to executing narrowly defined tasks within highly structured settings. They typically rely on extensive manual tuning, task-specific reward shaping, or large-scale expert demonstrations, which limit their flexibility and scalability. As a result, there remains a significant gap between current autonomous drone systems and truly intelligent agents that can understand abstract human intentions and complete complex tasks specified through natural language instructions.

\hfill
\begin{figure}[t]
    \flushright
    \includegraphics[width=\linewidth]{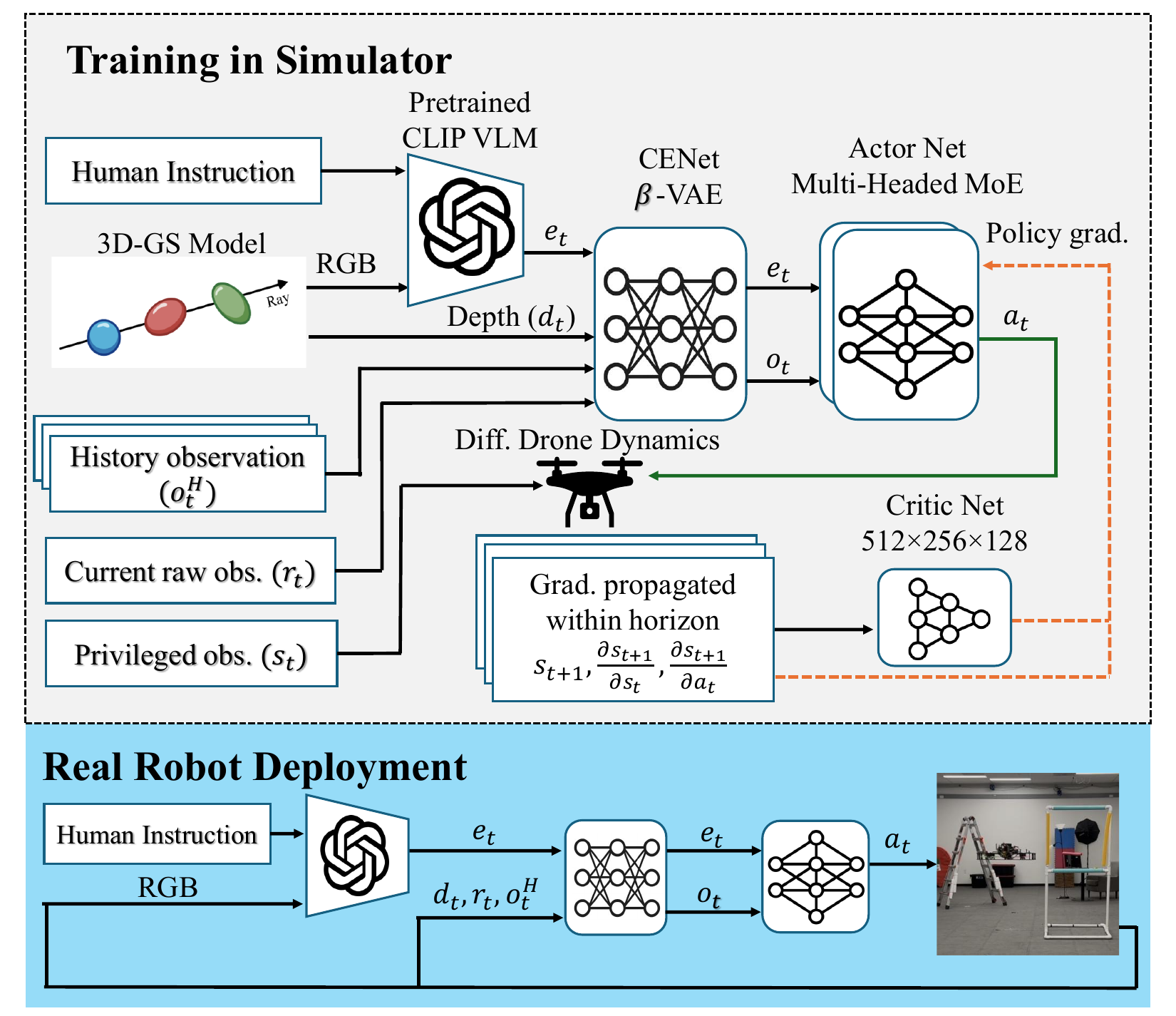} 
    \captionof{figure}{Our drone flight VLA model architecture.}
    \label{fig:model_struc}
\end{figure}

The emergence of large language models (LLMs)~\cite{mann2020language}, vision-language models (VLMs)~\cite{radford2021learning, zhai2023sigmoid}, has opened new possibilities for bridging the gap between natural language instructions and autonomous drone control. By leveraging the semantic understanding and reasoning capabilities of these large models~\cite{sun2023survey}, drones are now able to interpret high-level human commands and adapt their behaviors accordingly, enabling more flexible and intuitive human-robot interaction. 

Several recent works have explored the use of LLMs to direct drones in performing complex tasks and engaging in interactive behaviors~\cite{chen2023typefly, wang2025gsce}. However, these approaches typically adopt a layered architecture, where the LLM selects from a predefined set of task-specific skills or interacts with the system through API calls. Thus, the resulting gap between high-level decision-making and low-level control remains a critical bottleneck. Moreover, due to their size and computational demands, LLMs cannot be deployed onboard, requiring ground station support and reliable communication links—conditions that are often impractical in real-world deployments.

Vision-Language-Action (VLA) models~\cite{brohan2023rt, kim2024openvla, black2410pi0} offer a promising solution by enabling end-to-end policies that directly map natural language instructions and visual inputs to low-level actions. This unified architecture allows better use of pretrained model priors while avoiding the reliance on predefined skills or external APIs. Recent efforts have applied VLA models to drone flight tasks. RaceVLA~\cite{serpiva2025racevla} achieves impressive performance in competitive drone racing scenarios, but its applicability is limited to racing tasks and lacks generalization to more diverse navigation tasks. CognitiveDrone~\cite{lykov2025cognitivedrone} demonstrates embodied reasoning capabilities and instruction-grounded control, yet it is trained and evaluated entirely in simulation, with no evidence of sim-to-real transfer. In addition, both methods still rely on large-scale models with billions of parameters, making them unsuitable for real-time onboard deployment on computationally limited aerial platforms. 

% Vision-Language-Navigation (VLN)~\cite{liu2023aerialvln,saxena2025uav,cai2025flightgpt,chen2023vision} is a paradigm where agents follow natural language instructions to reach target locations, typically by predicting high-level actions like waypoints or navigation commands. Unlike VLA, VLN does not produce low-level control signals, and instead relies on separate planners or controllers for execution. This can introduce inconsistencies between language interpretation and control behavior, as errors can cumulate between modules. Our work adopts the VLA approach by directly predicting drone actions from visual input and language, avoiding such issues and enabling coherent end-to-end control.

% \textcolor{red}{Mac: Please exand the related work with Vision-Language-Navigation VLN papers \cite{liu2023aerialvln,saxena2025uav,cai2025flightgpt,chen2023vision} and any other related works.}
% \textcolor{blue}{Qianzhong: Added paragraph on VLN, emphasize the difference between VLA and why our work is VLA rather than VLN.}

A major barrier in training such models is the need for both high-fidelity visual environments and sample-efficient policy optimization. To address this, we leverage two recent advances: (i) 3D Gaussian Splatting (3DGS)~\cite{kerbl2023gaussiansplatting} enables high-quality photorealistic rendering at interactive rates, providing rich visual input without the need for costly mesh-based simulation or NeRF rendering; (ii) Differentiable Deep Reinforcement Learning (DiffRL)~\cite{xu2022accelerated}, a framework that uses differentiable physics simulation~\cite{freeman2021brax, howell2022dojo} to accelerate policy learning via gradient-based optimization. Built on this foundation, we further integrate a lightweight Mixture-of-Experts (MoE) action head, which improves generalization by routing computation across specialized sub-policies. Together, these components enable fast, high-fidelity, and sample-efficient training of VLA models, paving the way for real-time, onboard deployment of instruction-following drones. The key contributions of our approach are as follows:

\begin{itemize}
\item We propose a light-weighted drone flight VLA framework that operates entirely on the drone's onboard computing hardware.
\item Our VLA policy enables the drone to accomplish navigation tasks based on high-level natural language instructions, demonstrating generalization to novel tasks and adaptation to multiple environments.
\item Our policy uses a new multi-headed MoE action module, trained using 3DGS and DiffRL, achieving state-of-the-art performance in terms of sample efficiency and task success rate.
\end{itemize}

\section{Background} 
\label{sec:background}

% \subsection{Partially Observable Markov Decision Process (POMDP)}
% A POMDP is represented by the tuple $(S, O, A, P, r, \gamma)$. In this formulation, $S$ denotes the true state space, $O$ corresponds to the visual observation space, and $A$ is the action space of the robot. The transition dynamics are defined by $P: S \times A \rightarrow S$, and the reward function is given by $r: S \times A \rightarrow \mathbb{R}$. The discount factor $\gamma \in (0, 1]$ determines the relative importance of future rewards. The objective is to learn an optimal policy $\pi_\theta(a_t \mid o_t)$ that maximizes the expected cumulative reward $\mathbb{E}_{\pi} \left[ \sum_{t=0}^{\infty} \gamma^t r(s_t, a_t) \right]$.

\subsection{GRaD-Nav's Joint Simulation Pipeline with 3DGS and Differentiable Drone Dynamics}

\subsubsection{3D Gaussian Splatting}
3D Gaussian Splatting (3DGS) represents a scene using a set of anisotropic Gaussian primitives, each defined by position $\bm{\mu}_i$, covariance $\bm{\Sigma}_i$, color $\bm{c}_i$, and opacity $\alpha_i$~\cite{kerbl2023gaussiansplatting}. These Gaussians are projected onto a 2D plane to compute pixel colors:
\begin{equation}
\mathbf{C}(\mathbf{p}) = \sum_{i=1}^N \mathcal{N}(\mathbf{p}; \bm{\mu}_i, \bm{\Sigma}_i) \cdot T_i \cdot \bm{c}_i,
\end{equation}
where $T_i = \prod_{j=1}^{i-1} (1 - \alpha_j)$ accounts for transmittance and occlusion.

\subsubsection{Differentiable Quadrotor Dynamics}
\label{sec:background_simulation}
We derived from GRaD-Nav's~\cite{chen2025grad} PyTorch based differentiable quadrotor simulator to support gradient-based learning. The control input $\bm{u}_t = (\bm{\omega}_t^d, c_t)$ commands body rates and normalized thrust. The angular acceleration is:
\begin{equation}
\dot{\bm{\omega}} = \bm{I}^{-1}[\bm{K}_p(\bm{\omega}^d - \bm{\omega}) - \bm{K}_d\dot{\bm{\omega}} - \bm{\omega} \times (\bm{I}\bm{\omega})],
\end{equation}
and orientation $\bm{q} \in \mathbb{S}^3$ evolves as:
\begin{equation}
\bm{q}_{t+1} = \operatorname{norm}\left(\bm{q}_t + \frac{\Delta t}{2}\bm{q}_t \otimes [0\ \bm{\omega}]^\top\right).
\end{equation}
Linear acceleration is:
\begin{equation}
\bm{a} = \frac{1}{m} \bm{R}(\bm{q}) [0\ 0\ T]^\top + \bm{g}, \quad T = c T_{\max}.
\end{equation}

\subsubsection{Hybrid Simulation Pipeline}
At each simulation step, the drone pose $\bm{T} = (\bm{p}, \bm{q})$ is fed into the 3DGS renderer to generate first-person RGB observations. A precomputed point cloud also supports reward shaping, reference planning, and collision detection during DiffRL training.

\subsection{Mixture-of-Experts (MoE)}
The Mixture-of-Experts (MoE) architecture, initially proposed by Jacobs et al. \cite{jacobs1991adaptive} and Jordan \& Jacobs \cite{jordan1994hierarchical}, is a versatile model that employs a set of specialized expert networks, each focusing on different aspects of the input data. A central gating network determines which experts are activated for each input, dynamically selecting the most suitable ones. This design has been further refined by Shazeer et al. \cite{shazeer2017outrageously}, introducing a sparse MoE approach where only a subset of experts is utilized for each input. Our MoE implementation is derived from MENTOR \cite{huang2024mentor}:

\begin{equation}
F^{\text{MoE}}(\mathbf{x}) = \sum_{i=1}^{N} w_i(\mathbf{x}) \cdot \operatorname{FFN}_i(\mathbf{x}),
\end{equation}

\begin{equation}
w_i(\mathbf{x}) = \frac{\exp(h_i(\mathbf{x})) \cdot \mathbb{I}[i \in \mathcal{K}(\mathbf{x})]}{\sum_{j \in \mathcal{K}(\mathbf{x})} \exp(h_j(\mathbf{x}))},
\end{equation}

where $N$ is the total number of expert networks. $\operatorname{FFN}_i$ denotes the $i$-th expert, which is typically implemented as a feedforward neural network. The gating weight $w_i(\mathbf{x})$ determines the contribution of the $i$-th expert to the final output for input $\mathbf{x}$, and is computed by applying a softmax function over the scores $h_i(\mathbf{x})$ produced by a routing network. Only the top-$k$ experts (indexed by $\mathcal{K}(\mathbf{x})$) are selected for each input $\mathbf{x}$, and the remaining experts receive zero weight. The indicator function $\mathbb{I}[i \in \mathcal{K}(\mathbf{x})]$ ensures sparsity by masking out non-selected experts.

\section{Method}
We present a novel VLA-style drone navigation framework that can follow natural language instructions and conduct safe navigation in different environments. 

\subsection{Task Definition}
\label{sec:task_def}

We posit that enabling VLA-based drone flight requires two key capabilities: 

\emph{i. Multi-task generalization:}  
We train our VLA policy on a set of two-stage tasks within a single environment to promote generalization and zero-shot transfer to \textbf{untrained tasks}. Each task involves (1) selecting a correct direction from \textit{\{through, left, right, above\}} to pass a gate, and (2) identifying and flying to a target object among \textit{\{ladder, cart, monitor\}}. For example, the instruction ``GO THROUGH gate then FLY to LADDER base'' requires sequential spatial reasoning and object grounding. We construct 12 such tasks, training on 8 with guidance from a reference trajectory, and reserving 4 for zero-shot evaluation. All directions and objects are evenly represented in training.

\emph{ii. Multi-environment adaptation:}  
We also train the policy on a smaller task set across two distinct 3DGS environments, each with different gate placements and distractors. These single-stage tasks require directional decisions (e.g., ``FLY past the LEFT side of the gate'') based on vision-language input. Importantly, we do not signal environment changes—robust adaptation emerges from the VLM-based visual grounding.

Table~\ref{table:task_instruc} summarizes the task instructions. For each trained task, we generate a reference trajectory using $n=4$ key waypoints defined on the 3DGS point cloud and connected via $A^*$ planning. Further details are provided in Section~\ref{sec:drone_rl}.

\subsection{Model Architecture}
An overview of our model architecture is demonstrated in Figure~\ref{fig:model_struc}.
\subsubsection{Vision-Language Model (VLM)}
\label{sec:method_VLM}
We use a pretrained CLIP model \cite{radford2021learning} for high-level scene understanding and instruction matching. Natural language instructions and first-person RGB images are encoded by CLIP’s text and visual encoders, while a trainable linear layer fuses and downsamples the embeddings into a common 512-D feature vector $\mathbf{e}_t$, used as input to the policy. To meet onboard Orin Nano constraints, CLIP runs asynchronously with the policy and updates $\mathbf{e}_t$ every 10 steps. This provides the policy with semantically rich representations for decision making and control.

\subsubsection{Policy Network}
\label{sec:method_nn}
We employ a MoE policy network comprising two expert subnetworks. At each time step, the router activates the top-$k$ experts with $k{=}2$, ensuring that all experts are utilized while maintaining sparse computation. The number of experts is intentionally kept small to enable efficient inference and real-time deployment on resource-constrained onboard hardware.
 Each expert $\pi_{\theta}(\mathbf{a}_{t+1}|\mathbf{o}_t, \mathbf{e}_t)$ is a multi-headed network parameterized by $\theta$, which takes the VLM feature vector $\mathbf{e}_t$ and the observation $\mathbf{o}_t \in \mathbb{R}^{72}$ as input. Each head of the single expert is a 3-layer MLP and each layer has 512, 256, 128 neurons, respectively. The processed VLM feature and observation are fused and generate action $\mathbf{a}_{t+1}$ using another 3-layer MLP, each layer has 256, 128, 64 neurons, respectively. The observation is defined as following:
\begin{equation}
    \mathbf{o}_t = \begin{bmatrix} \bm{d}_t & \bm{z}_t & \mathbf{r}_t \end{bmatrix}^{T},
\end{equation}
where $\bm{d}_t \in \mathbb{R}^{32}$ is the minimum-pooled depth image that used for collision avoidance; $\bm{z}_t \in \mathbb{R}^{24}$ is the output of the context estimator network (will be introduced in Section~\ref{method:cenet}), which is used for mitigating sim-to-real gap and improving policy robustness; $\mathbf{r}_t$ is the raw dynamics observation from drone differentiable simulator and can be defined as:
\begin{equation}
    \mathbf{r}_t = \begin{bmatrix} \bm{h}_t & \bm{q}_t & \bm{v}_t & \bm{a}_t & \bm{a}_{t-1} \end{bmatrix}^{T},
\end{equation}
where $\bm{h}_t$, $\bm{q}_t$, $\bm{v}_t$ are drone body's height, quaternion, and linear velocity, respectively, and $\bm{a}_t$ and $\bm{a}_{t-1}$ are the current action and previous action. We do \textbf{not} need to explicitly estimate the drone's x-y positions. Aligned with our simulation setting Section~\ref{sec:background_simulation}, the action $\bm{a}_t = (\bm{\omega}^d_t, c_t)$ is a 4-dimensional vector, including body rate $\bm{\omega}^d_t \in \mathbb{R}^3$ and normalized thrust $c_t \in \mathbb{R}$. It is important to note that we do \textbf{not} implement a standalone convolution neural network (CNN) for RGBD visual perception. Instead, we directly utilize the VLM feature vector $\mathbf{e}_t$ as a high-level representation of visual semantics. This design choice is primarily motivated by the need to optimize inference speed during real-world deployment on the robot's onboard computer. Additionally, the depth feature vector $\bm{d}_t$ is derived via a simple minimum pooling operation and is used exclusively for low-level collision avoidance, rather than for high-level perceptual reasoning.

\subsubsection{Value Network}
The value network $V_{\phi}(\mathbf{s}_t, \mathbf{e}_t)$ shares the same multi-headed structure as policy network $\pi_{\theta}$, parameterized by $\phi$ and is used to estimate the state value. Except for the observation $\mathbf{o}_t$, the privileged observation $\mathbf{s}_t \in \mathbb{R}^{74}$ used by the value network also has access to body's x-y positions $\bm{p}_t \in \mathbb{R}^2$ defined as:
\begin{equation}
    \mathbf{s}_t = \begin{bmatrix} \mathbf{o}_t &  \bm{p}_t  \end{bmatrix}^{T}.
\end{equation}
The value function is not needed at runtime, hence the access to privileged information from the simulator is not a practical limitation.

\subsubsection{Context Estimator Network}
\label{method:cenet}
Similar to GRaD-Nav~\cite{chen2025grad}, we incorporated a $\beta$-variational autoencoder ($\beta$-VAE) \cite{higgins2017beta} based CENet~\cite{nahrendra2023dreamwaq} with visual perception data \re{to mitigate sim-to-real gap}. CENet is designed for encoding the drone's surrounding environment, especially the spatial relationship with obstacles, to a latent vector $\mathbf{z}_t$ to enable runtime adaptivity to the environment. We used a history observation of the last 5 time steps as CENet's input.

\subsection{Training Procedure}

\label{sec:drone_rl}
Derived from GRad-Nav~\cite{chen2025grad},
% \textcolor{red}{Mac: Might have to change wording for double blind?}, \textcolor{blue}{Qianzhong: reword to cut the relationship} 
our DiffRL training procedure centers on a reward function designed for smooth, stable dynamic control and efficient, safe navigation, as detailed in Table~\ref{table:reward_function}. This function, $r_t(\mathbf{s}_t, \mathbf{a}_t) = \sum r_i w_i$, is intentionally kept simple to enhance transferability across different environments and agents. 
% To foster a robust policy and improving sim-to-real transitions, we employ \textbf{domain randomization} on critical system dynamics parameters, with details listed in Table~\ref{table:domain_randomization}. 
For guiding the drone along desired trajectories, as described in Table~\ref{table:reward_function}, we incorporate several reward terms. A \textbf{waypoint reward} $r_{\text{wp}} = \left( e^{-{\|\bm{p} - \bm{w}_{\text{next}}\|^2 }} \right)$ encourages the drone (at position $\bm{p}$) to approach the next waypoint $\bm{w}_{\text{next}}$ (i.e., the closest waypoint with $x_{\text{wp}} > x_{\text{drone}}$) on a precomputed reference trajectory. To further ensure adherence to this path, a \textbf{reference trajectory tracking reward} $r_{\text{traj}} = \left\| \frac{\bm{v}}{\|\bm{v}\|} - \frac{\bm{v}_{\text{des}}}{\|\bm{v}_{\text{des}}\|} \right\|$ penalizes deviations between the drone's normalized velocity $\frac{\bm{v}}{\|\bm{v}\|}$ and the desired velocity direction $\frac{\bm{v}_{\text{des}}}{\|\bm{v}_{\text{des}}\|}$ from the reference trajectory. Safe navigation is promoted by an \textbf{obstacle avoidance reward} $r_{\text{obs}} = \bm{d}_{\text{obs}}$ when the minimum distance to the nearest obstacle $\bm{d}_{\text{obs}}$ within the drone's field of view is less than a predefined threshold $\bm{d}_{\text{th}}$ (set to 0.5m in our method). The specific weighting factors $\omega_i$ for these components are listed in Table~\ref{table:reward_function}.

To ensure balanced task exposure during training, we initialize each agent in the batch with a task sampled uniformly at random. Furthermore, upon each environment reset, agents are reassigned new tasks, also randomly selected from a uniform distribution. This strategy dynamically redistributes task assignments across the batch throughout training, thereby mitigating potential overfitting to specific tasks. For the task generalization experiment, we train the policy within a single environment for 1000 epochs, which requires approximately 8 hours on a single Nvidia RTX 4090 GPU. For the environment adaptation experiment, we alternate training between two distinct environments represented by 3DGS models. Specifically, the policy is trained for 100 epochs in one environment before switching to the other, with this cycle repeated until 1500 epochs are completed in each environment. 

% The total training duration under this multi-environment setting is approximately 14 hours on the same desktop mentioned above. Due to the additional computational overhead introduced by VLM inference during forward simulation, our approach incurs a 5\% wall-clock time increase during training compared to the original GRaD-Nav framework.

% \textcolor{red}{Mac: Again, reword to avoid double blind violation.} \textcolor{blue}{Qianzhong: removed this paragraph, not necessary}

\begin{table}[h]
    \centering
    \small
    \renewcommand{\arraystretch}{1.3}
    \setlength{\tabcolsep}{3pt}
    \caption{Reward terms and weights. Early termination occurs if (i) height $>3$m, (ii) linear velocity $>20$ m/s, or (iii) the drone is out of bounds by $>3$m ($\bm{x}_{o.b.}, \bm{y}_{o.b.} \geq 3$). Here, $\bm{q}_0$ is the initial quaternion, $\bm{h}_{\text{target}}$ the target/hover height, $\hat{\bm{y}}_{\text{yaw}}$ the normalized heading, $\bm{d}_{\text{wp}}$ the distance to the next waypoint, $\bm{d}_{\text{obst}}$ the closest obstacle distance in FOV, $\bm{x}_{o.b.}, \bm{y}_{o.b.}$ the distances to map boundaries, and $\bm{v}_{\text{des}}$ the desired velocity from the reference trajectory.}
    \label{table:reward_function}
    \begin{tabular}{l c c}
        \toprule
        \textbf{Reward} & \textbf{Equation ($r_i$)} & \textbf{Weight ($w_i$)} \\
        \hline
        \multicolumn{3}{c}{\textbf{Safe Control Rewards}} \\
        \hdashline
        Survival & $\notin \{\text{early terminations}\}$ & $8.0$ \\
        Linear velocity & $\|\bm{v}\|^2$ & $-0.5$ \\
        Pose & $\|\bm{q} - \bm{q}_0\|$ & $-0.5$ \\
        Height & $(\bm{h} - \bm{h}_{\text{target}})^2$ & $-2.0$ \\
        Action & $\|\bm{a}_t\|^2$ & $-1.0$ \\
        Action rate & $\|\bm{a}_t - \bm{a}_{t-1}\|^2$ & $-1.0$ \\
        Smoothness & $\|\bm{a}_t - 2\bm{a}_{t-1} + \bm{a}_{t-2}\|^2$ & $-1.0$ \\
        \hline
        \multicolumn{3}{c}{\textbf{Efficient Navigation Rewards}} \\
        \hdashline
        Yaw alignment & $\frac{\bm{v}_{xy}}{\|\bm{v}_{xy}\|} \cdot \hat{\bm{y}}_{\text{yaw}}$ & $0.25$ \\
        Waypoint & $\exp(-\bm{d}_{\text{wp}})$ & $2.0$ \\
        Obstacle avoidance & $\bm{d}_{\text{obst}}$ & $1.0$ \\
        Out-of-map & $\bm{x}_{\text{o.b.}}^2 + \bm{y}_{\text{o.b.}}^2$ & $-2.0$ \\
        Ref. traj. tracking & $\|\frac{\bm{v}}{\|\bm{v}\|} - \frac{\bm{v}_{\text{des}}}{\|\bm{v}_{\text{ref}}\|}\|$ & $-2.0$ \\
        \bottomrule
    \end{tabular}
\end{table}

\begin{figure}[h]
    \centering
    \includegraphics[width=0.8\linewidth]{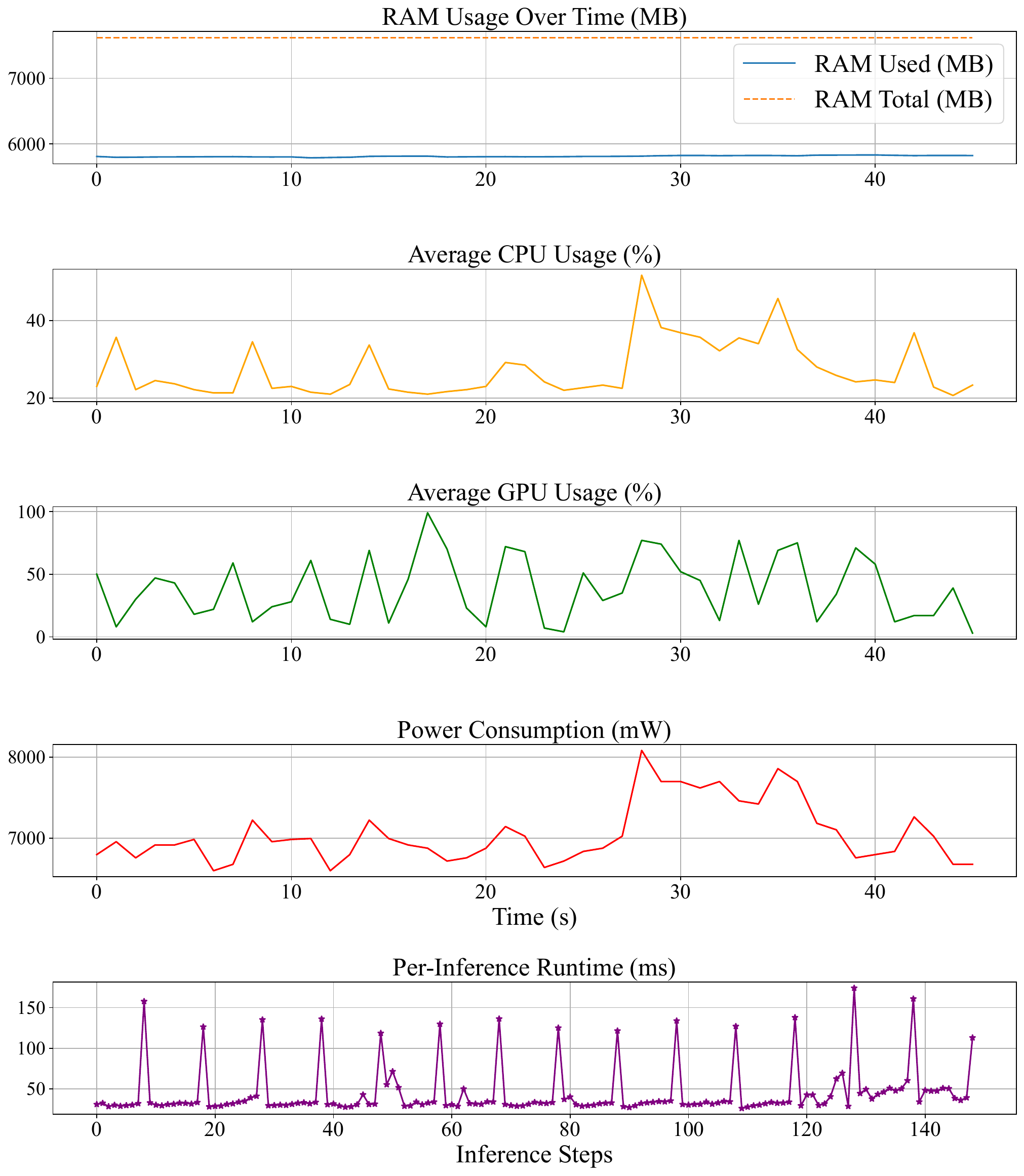}
    \caption{\re{Runtime profiling of real-time inference on NVIDIA Jetson Orin Nano.}}
    \label{fig:runtime_spec}
\end{figure}

\re{
\subsection{Onboard Inference Profile}
We profile the onboard inference behavior of our model running on an NVIDIA Jetson Orin Nano at Figure~\ref{fig:runtime_spec}. The data was collected during an actual flight experiment, powered by a fully charged 4S (14.8\,V) LiPo battery. The full model during inference time consists of three components: a \texttt{clip-vit-base-patch32} vision encoder (150M parameters), a CENet encoder (0.5M parameters), and a mixture-of-experts (MoE) policy network (1M parameters). During execution, our memory usage remains stable at approximately 6\,GB. We collect system metrics including RAM usage, CPU load, GPU utilization, and power consumption throughout the execution. For runtime profiling, in the nominal steps without CLIP, the policy executes quickly at around 30\,ms per step. When CLIP is invoked (every 10 steps), the runtime surges to over 120\,ms. These periodic latency spikes align precisely with CLIP execution and are clearly visible in the runtime plot.

}

\section{Experimental Results}

In this section, we aim to address the following research questions through empirical evaluation:
\begin{enumerate}
\item What is the functional contribution of the VLM within our VLA drone flight framework?
\item In what ways does the MoE architecture facilitate multi-task, multi-environment \re{adaptation} and mitigate catastrophic forgetting?
\item Is the use of DiffRL essential to the effectiveness of our proposed approach?
\end{enumerate}

We conduct two experiments that focus on new task generalization and multi-environment adaptation in both simulation and on a real drone, respectively. For simulation, the drone's initial positions are randomized in a cube with side length of 1m, centered at (0,0,1.4)m; its initial poses are also randomized, $\phi, \theta, \psi \in [-0.25,0.25].$ For real hardware experiments, our drone is mounted with a Pixracer low-level flight controller and an Intel Realsense D435 camera. All the models are deployed and running onboard using a NVIDIA Jetson Orin Nano companion computer, with overall control frequency at 25 Hz. 
\subsection{Multi Tasks Generalization Experiment}
\label{sec:exp_multi_task}

In this experiment, we evaluate our VLA-based drone framework on multiple long-horizon tasks, focusing on generalization to \textbf{unseen tasks}. To assess the VLM’s contribution, we conduct an ablation study by replacing it with one-hot encoded task instructions. We also compare against a Proximal Policy Optimization (PPO)~\cite{schulman2017proximal} baseline using the same VLM to highlight the sample efficiency benefits of DiffRL. As defined in Section~\ref{sec:task_def}, our tasks involve two-stage instructions requiring spatial reasoning and object recognition. Each task pairs a natural language instruction with a reference trajectory. We train on 8 tasks and evaluate on 4 held-out tasks, using identical training conditions and simulation steps across all methods.

\begin{figure}[h!]
    \centering
    \includegraphics[width=\linewidth]{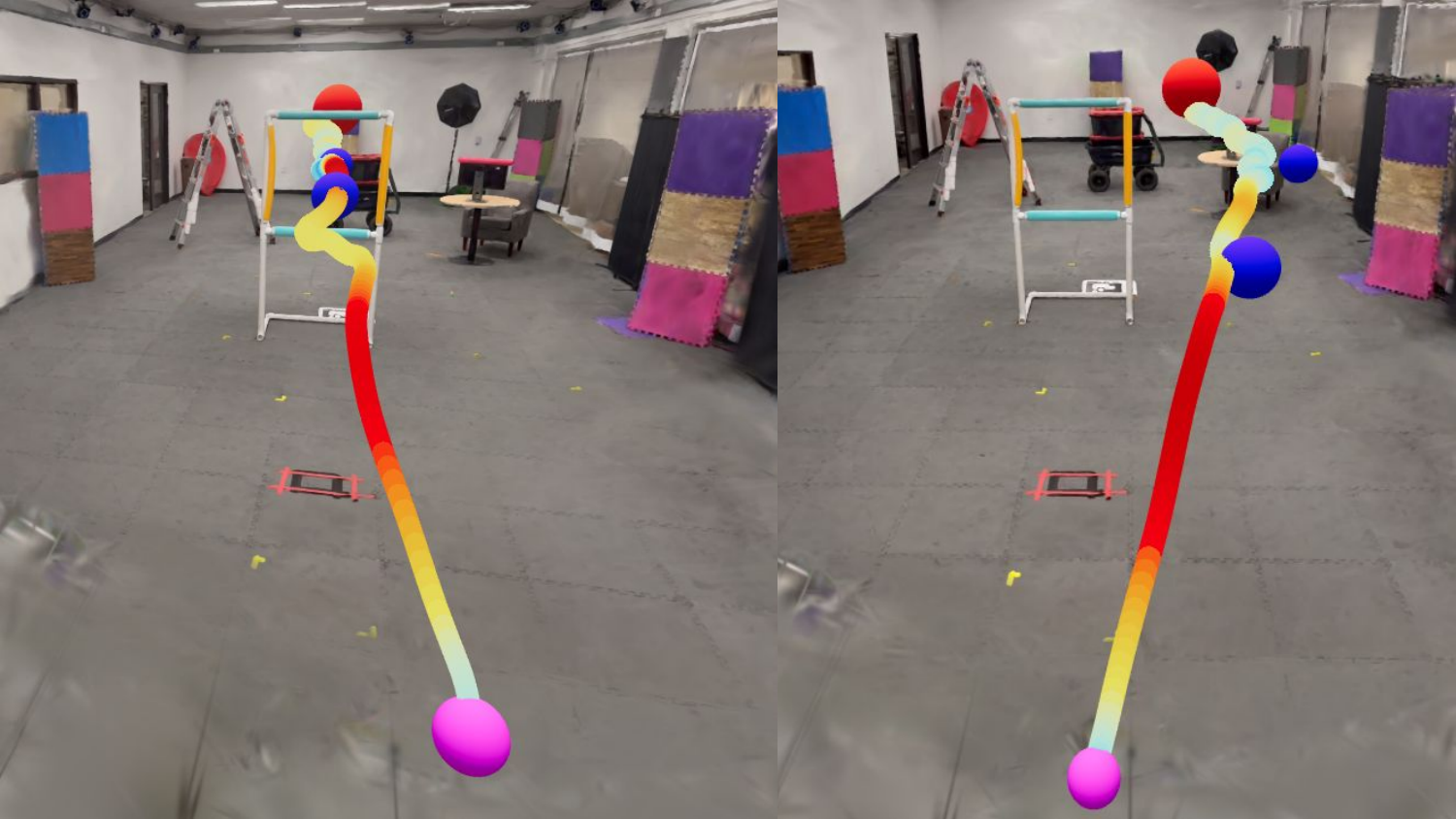}
    \caption{Example trajectories of untrained long horizon tasks. The instructions are ``GO THROUGH gate then STOP over CART" (left) and ``FLY past the RIGHT side of the gate then STOP over MONITOR" (right). }
    \label{fig:multi_task_sim}
\end{figure}

We report success rates (SR) for each stage and the overall task success, in both simulation and real drone tests. Stage 2 SR is only evaluated if Stage 1 succeeds. Beyond avoiding early termination and collisions (see Table~\ref{table:reward_function}), Stage 1 requires navigating through a gate in the correct direction, while Stage 2 involves identifying and approaching the correct object. A Stage 2 trial is considered successful if the drone ends up closer to the target object than to either distractor:
\begin{equation}
\|\mathbf{p} - \mathbf{p}_\text{t}\| < \min\left( \|\mathbf{p} - \mathbf{p}_1\|,\ \|\mathbf{p} - \mathbf{p}_2\| \right)
\end{equation}
where $\mathbf{p}$ is the drone’s final position, and $\mathbf{p}_\text{t}$, $\mathbf{p}_1$, and $\mathbf{p}_2$ are the positions of the target and distractor objects. Figure~\ref{fig:multi_task_sim} shows examples of successful generalization in simulation.

The results are summarized in Table~\ref{table:multi_task_sim} and Table~\ref{table:multi_task_real} for simulation and real-world experiments, respectively.

\begin{table}[t]
\centering
\scriptsize
    \caption{Simulation results of multi task generalization experiment. For 8 trained tasks and 4 untrained tasks, the evaluation is conducted on 6 trials for each task, with the average rewards and success rates (SR) reported. \re{``Proposed, LR" denotes the ablation study of training the policy in imperfect 3DGS.}}
    \label{table:multi_task_sim}
    \begin{tabular}{ccccc}
    \toprule
    \multirow{2}{*}{\textbf{Method}} & \multicolumn{4}{c}{\textbf{Evaluation Results (Trained \textbar\ Untrained)}}                                        \\
                            & Reward              & Stage 1 SR             & Stage 2 SR            & Overall SR            \\ \midrule
    PPO               & 1272.8 \textbar\ 1168.3          & 0/24 \textbar\ 0/12            & 0/24 \textbar\ N/A            & 0/24 \textbar\ 0/12           \\
    w/o VLM                 & 3725.7 \textbar\ 3246.5          & 3/24 \textbar\ 1/12            & 1/3 \textbar\ 0/1             & 1/24 \textbar\ 0/12           \\
    \re{Proposed, LR}                & \re{4873.3} \textbar\ \re{4322.8} & \re{22/24} \textbar\ \re{\textbf{10/12}} & \re{14/22} \textbar\ \re{6/10} & \re{14/24} \textbar\ \re{6/12} \\
    Proposed                & \textbf{5068.8} \textbar\ \textbf{4447.3} & \textbf{24/24} \textbar\ \textbf{10/12} & \textbf{20/24} \textbar\ \textbf{9/10} & \textbf{20/24} \textbar\ \textbf{9/12} \\ \bottomrule
    \end{tabular}
\end{table}

\begin{table}[h!]
\centering
\small
    \caption{Real hardware experiment results of multi-task generalization experiment. For 8 trained tasks and 4 untrained tasks, the evaluation is conducted on 3 trials for each task, with the success rates (SR) reported. \re{``Proposed, LR" denotes the ablation study of training the policy in imperfect 3DGS.}}
    \label{table:multi_task_real}
    \begin{tabular}{cccc}
    \toprule
    \multirow{2}{*}{\textbf{Method}} & \multicolumn{3}{c}{\textbf{Success Rate (Trained \textbar\ Untrained)}}               \\
                            & Stage 1               & Stage 2              & Overall               \\ \midrule
    PPO               & 0/24 \textbar\ 0/12           & 0/24 \textbar\ N/A           & 0/24 \textbar\ 0/12           \\
    w/o VLM                 & 1/24 \textbar\ 0/12           & 0/1 \textbar\ N/A            & 0/24 \textbar\ 0/12           \\
    \re{Proposed, LR}                & \re{16/24 \textbar\ 7/12} & \re{8/16 \textbar\ 2/7} & \re{9/24 \textbar\ 2/12} \\
    Proposed                & \textbf{21/24 \textbar\ 9/12} & \textbf{16/21 \textbar\ 6/9} & \textbf{16/24 \textbar\ 6/12} \\ \bottomrule
    \end{tabular}
\end{table}

% The experimental results reveal that (i) nominal model-free RL methods, such as PPO, fail to converge to an effective policy within $1 \times 10^{7}$ simulation steps when applied to our VLA-based drone flight tasks involving multi-task generalization. (ii) the VLM module proves to be essential for the success of our framework. Beyond serving as a text encoder, the VLM plays a pivotal role in grounding natural language instructions to the visual context of the environment, thereby enabling the drone to interpret and carry out complex tasks. (iii) our proposed method achieves high overall success rates of 83\% and 75\% on trained and unseen tasks, respectively, in simulation. When deployed on real hardware, the method maintains robust performance, with success rates equal to or exceeding 50\% for both trained and unseen tasks—demonstrating effective sim-to-real transfer. (iv), Stage 2 experiences a more pronounced decline in success rate than Stage 1 during sim-to-real deployment. This degradation is likely due to a combination of factors, including the increased challenge of reliable object detection, the requirement for more precise control when approaching target objects, and the accumulation of compounding errors.

The results show that: (i) standard model-free RL (e.g., PPO) fails to learn effective policies for our VLA-based multi-task drone flight within $1 \times 10^{7}$ steps; (ii) the VLM module is critical—not just as a text encoder, but for grounding instructions to visual context, enabling task understanding; (iii) our method achieves 83\%/75\% success on trained/unseen tasks in simulation, and over 50\% on both in real-world tests, indicating strong sim-to-real transfer; (iv) performance drops more in Stage 2 than Stage 1 on hardware, likely due to challenges in detection, fine control, and error accumulation.

\re{
We conduct an ablation study with a degraded 3DGS map to test policy robustness under imperfect visual perception. The degraded 3DGS, trained with only one-fifth of the steps of the high-quality model, produces blurrier renderings with distorted objects (Figure~\ref{fig:3dgs_compare}). Results are reported in Table~\ref{table:multi_task_sim} and Table~\ref{table:multi_task_real} as ``Proposed, LR.'' While overall performance drops, the policy still maintains a reasonable success rate. The performance gap between simulation and real hardware grows under degraded visuals, highlighting the importance of rendering quality for sim-to-real transfer. In particular, Stage~2 suffers more than Stage~1: Stage~1 only requires selecting the correct gate direction, which tolerates low-quality visuals, whereas Stage~2 demands recognizing and approaching the correct object. These findings show that although 3DGS quality is crucial, our method remains robust to degraded conditions.

\begin{figure}[h]
    \centering
    \includegraphics[width=\linewidth]{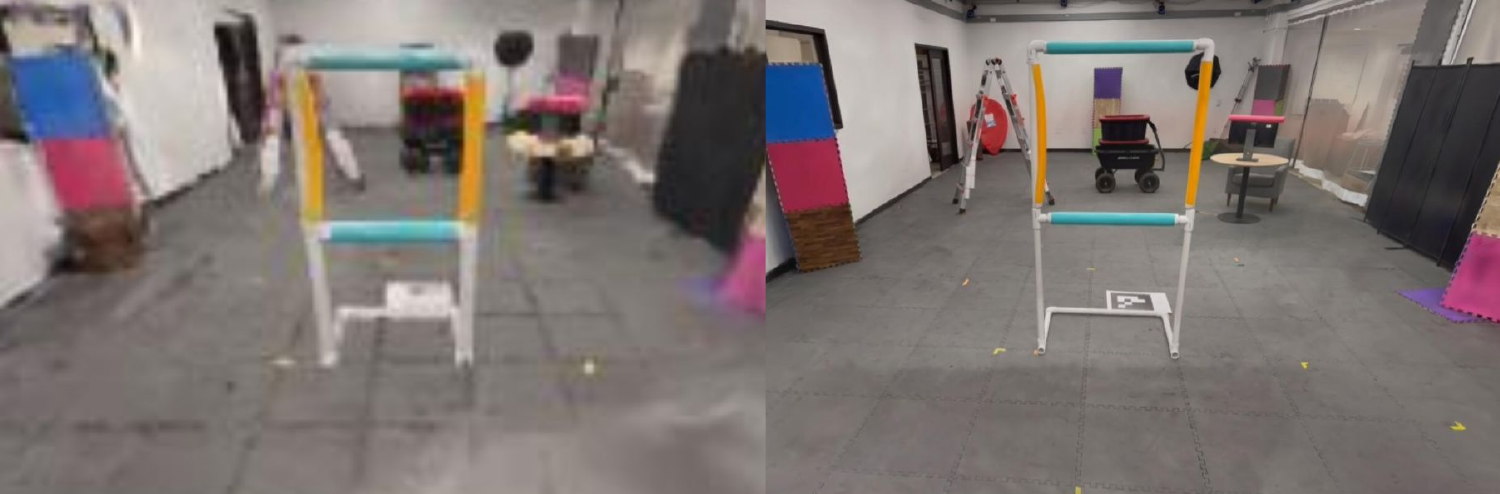}
    \caption{\re{Comparison of imperfect (left) and high quality (right) 3DGS maps' rendering effect.}}
    \label{fig:3dgs_compare}
\end{figure}
}

\subsection{Multi Environment Adaptation Experiment}
\begin{figure*}[t]
    \centering
    \includegraphics[width=\linewidth]{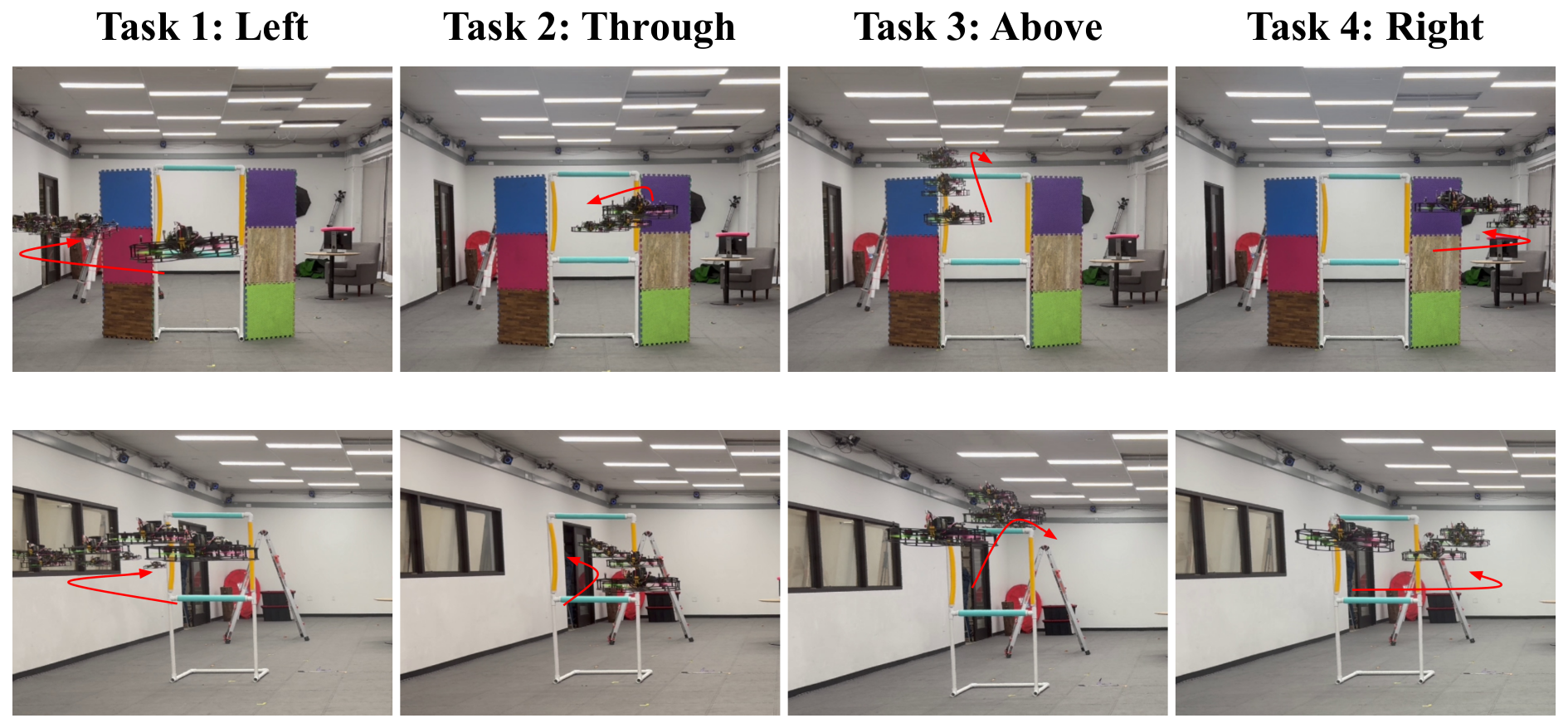}
    \caption{Demonstration of multi-environment adaptation in real-world experiments using video frame overlay visualization. The top row shows the drone flying \textit{to the left of}, \textit{through}, \textit{above}, and \textit{to the right of} the gate in the \textit{middle-gate} environment. The bottom row shows the drone executing the same directional tasks in the \textit{left-gate} environment. Red arrowed curves illustrate the approximate flight trajectories. The learned policy demonstrates robust \re{adaptation} to varying environments, adapting to changes in gate positions and the presence of distractor objects.}
    \label{fig:multi_env_combo}
\end{figure*}

\label{sec:exp_multi_env}
In this experiment, we evaluate our drone VLA framework’s ability to follow human instructions and perform multiple tasks across \textbf{multiple environments}, focusing on the role of the MoE architecture in enabling multi-task, multi-environment generalization while reducing catastrophic forgetting. We compare the MoE actor against two MLP-based ablations: (i) a \textit{single expert} (SE) baseline that uses the same architecture as an individual MoE expert, and (ii) a \textit{large network} (LN) baseline that concatenates two such experts (matching the MoE’s parameter count with $top\_k=2$).

\begin{table}[t]
\centering
\scriptsize
    \caption{Simulation results of the multi-environment adaptation experiment are presented. For each of the 4 tasks in the 2 different surrounding environments, evaluation is conducted over 6 trials per task. The reported metrics include the average evaluation rewards and success rates.}
    \label{table:multi_env_sim}
    \begin{tabular}{cccccc}
    \toprule
    \multirow{2}{*}{\textbf{Method}} & \multirow{2}{*}{\textbf{Reward}} & \multicolumn{4}{c}{\textbf{Success Rate (Left \textbar\ Mid.)}}                                    \\
                            &                               & Through            & Right              & Left               & Over               \\
    \midrule
    % PPO                     & 573.2                        & 0/6 \textbar\ 0/6          & 0/6 \textbar\ 0/6          & 0/6 \textbar\ 0/6          & 0/6 \textbar\ 0/6          \\
    % w/o VLM                 & 2185.6                        & 0/6 \textbar\ 0/6          & 1/6 \textbar\ 0/6          & 0/6 \textbar\ 1/6          & 1/6 \textbar\ 1/6          \\
    w/o MoE, SE             & 3566.7                        & 5/6 \textbar\ 1/6          & 5/6 \textbar\ 0/6          & 5/6 \textbar\ 3/6          & 4/6 \textbar\ 2/6          \\
    w/o MoE, LN             & 3764.6                        & \textbf{6/6} \textbar\ 2/6 & \textbf{6/6} \textbar\ 1/6 & 5/6 \textbar\ 3/6          & \textbf{5/6} \textbar\ 2/6 \\
    Proposed                & \textbf{4169.5}               & \textbf{6/6} \textbar\ \textbf{5/6} & 5/6 \textbar\ \textbf{4/6}          & \textbf{6/6} \textbar\ \textbf{4/6} & \textbf{5/6} \textbar\ \textbf{4/6} \\
    \bottomrule
    \end{tabular}
\end{table}

\begin{table}[h!]
\centering
\small
    \caption{Real hardware experiment results of the multi-environment adaptation experiment. For each of the 4 tasks across 2 different surrounding environments, evaluation is performed over 3 trials per task, with the success rates reported.}
    \label{table:multi_env_real}
    \begin{tabular}{ccccc}
    \toprule
    \multirow{2}{*}{\textbf{Method}} & \multicolumn{4}{c}{\textbf{Success Rate (Left \textbar\ Mid.)}}                                    \\
                            & Through            & Right              & Left               & Over               \\ 
    \midrule
    w/o MoE, SE                 & \textbf{3/3} \textbar\ 0/3 & \textbf{3/3} \textbar\ 0/3 & 1/3 \textbar\ 1/3          & \textbf{2/3} \textbar\ 0/3 \\
    w/o MoE, LN       & 2/3 \textbar\ 0/3          & \textbf{3/3} \textbar\ 1/3 & 1/3 \textbar\ 2/3          & \textbf{2/3} \textbar\ 0/3 \\
    Proposed                & \textbf{3/3} \textbar\ \textbf{2/3} & \textbf{3/3} \textbar\ \textbf{2/3} & \textbf{2/3} \textbar\ \textbf{3/3} & \textbf{2/3} \textbar\ \textbf{1/3} \\
    \bottomrule
    \end{tabular}
\end{table}

The tasks used in this experiment are defined in Section~\ref{sec:task_def}, focusing on spatial reasoning and correct gate traversal, i.e. the first stage of the tasks introduced in Section~\ref{sec:exp_multi_task}. We evaluate performance in both simulation and real-world deployments, reporting training reward, evaluation reward, and task-specific success rates (SR) across environments. The success criterion is identical to that of Stage 1 as defined in Section~\ref{sec:exp_multi_task}: the drone must traverse the designated gate in the correct direction as specified by the instruction, without incurring collisions or early termination. 
% The training hyperparameters are provided in Table~\ref{table:hyper-param}. 
The simulation and real hardware experiments results are demonstrated in Table~\ref{table:multi_env_sim} and Table~\ref{table:multi_env_real}, respectively. Figure~\ref{fig:multi_env_combo} shows real-world hardware demonstrations of the drone executing tasks in multiple environments.

The proposed MoE policy \re{adaptation} well across environments. In the \textit{left gate} setting, all methods—SE, LN, and MoE—perform comparably. However, in the \textit{mid gate} environment, SE and LN degrade sharply (SE: 25\% sim / 8\% real, LN: 33\% sim / 25\% real), while MoE maintains 70\% in simulation and 67\% on hardware. This is due to catastrophic forgetting in SE and LN, which overfit to the last trained (left gate) task. In contrast, MoE avoids forgetting by dynamically routing through experts, preserving environment-specific knowledge and enabling robust performance across tasks.

\begin{figure}[h]
    \centering
    \includegraphics[width=\linewidth]{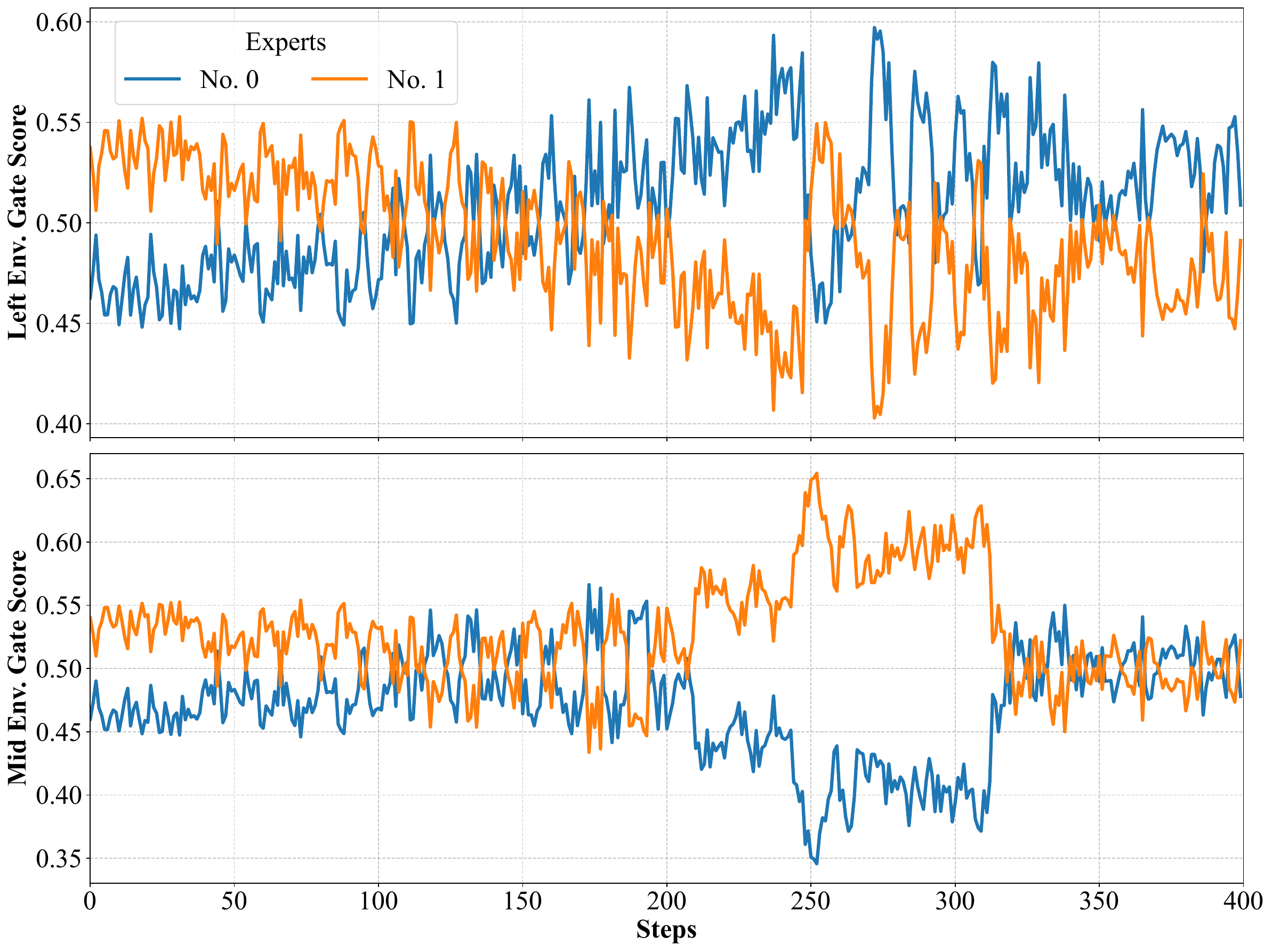}
    \caption{Experts' usage intensity when executing the same task (``GO THROUGH gate") at different surrounding environments (top-\textit{left gate}, bottom-\textit{middle gate}).}
    \label{fig:moe_spec}
\end{figure}

Figure~\ref{fig:moe_spec} illustrates the expert utilization patterns during the execution of the same task (``GO THROUGH gate") under different surrounding environments. The results demonstrate that the MoE architecture adaptively allocates expert resources according to the specific demands of each environment. In both plots, a noticeable shift in expert activation occurs between time steps 200 and 300, which corresponds to the phase when the drone approaches the gate. This transition indicates that the gating network dynamically adjusts expert weights in response to changing environmental context and task phase. These observations further support the effectiveness of the MoE architecture in enabling efficient resource allocation and facilitating robust \re{adaptation} across multiple tasks and environments.

\subsection{Task Shift Experiment}
\label{sec:exp_task_shift}
To further validate and analyze the functional contribution of the VLM within our VLA-based drone flight framework, we design an experiment in which the task instruction is modified during the execution of a long-horizon task. This experiment aims to assess the VLM's ability to re-ground a new instruction within the current visual context and evaluate whether the learned policy can successfully adapt mid-flight to a new task objective.

\begin{figure}[h]
    \centering
    \includegraphics[width=\linewidth]{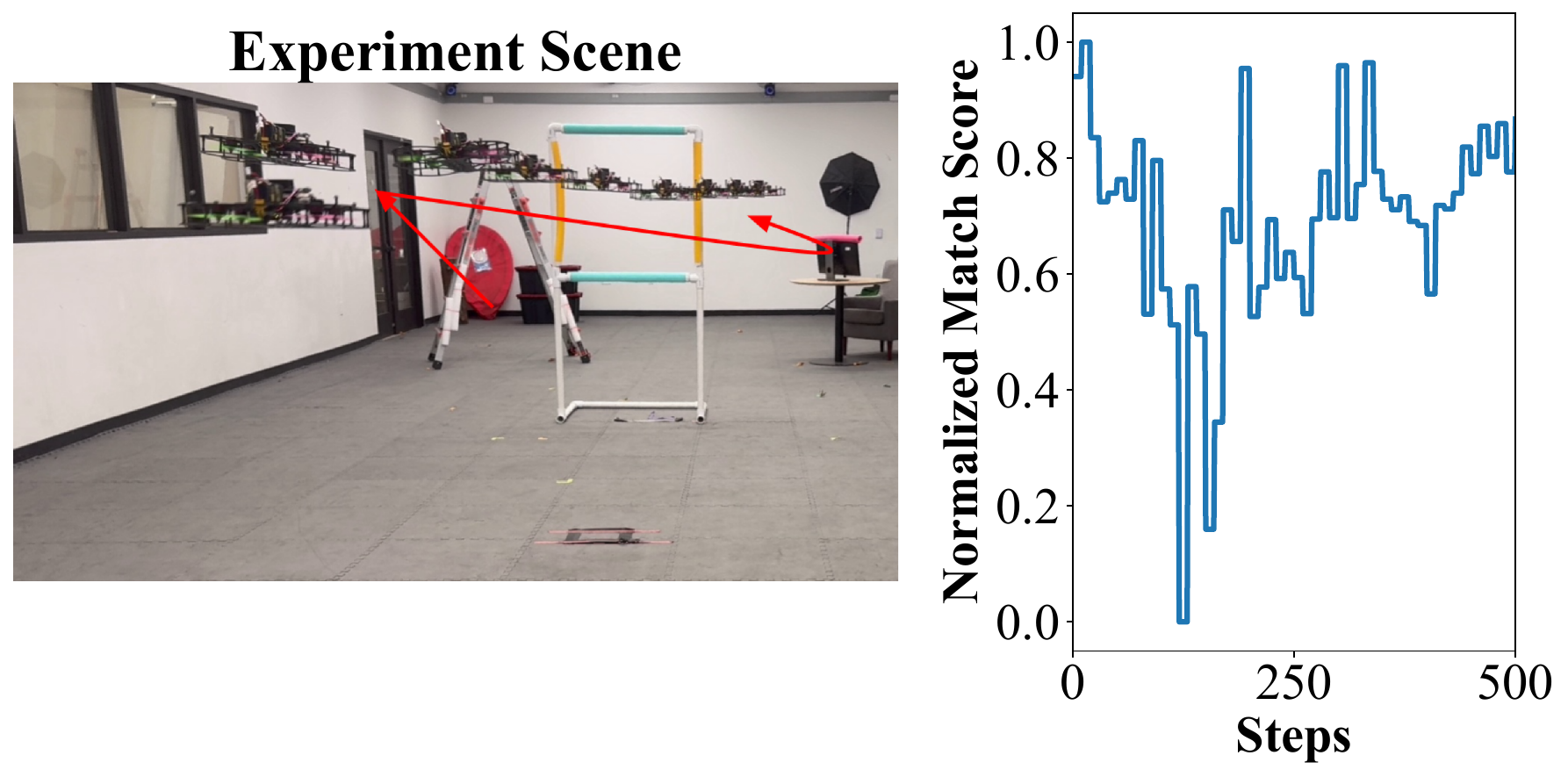}
    \caption{\re{Task-switching with instruction change at step 100. Left: drone flying past the gate. Right: cosine similarity between text and visual embeddings, showing VLM re-grounding.}}
    \label{fig:task_shift}
\end{figure}

During execution, we record the cosine similarity between the VLM's text and visual embeddings every 10 time steps, matching the update frequency of the VLM module. The cosine similarity is computed as:
\begin{equation}
\text{cosine\_similarity}(\bm{e}_\text{tex}, \bm{e}_\text{vis}) = \frac{\bm{e}_\text{tex} \cdot \bm{e}_\text{vis}}{\|\bm{e}_\text{tex}\| \|\bm{e}_\text{vis}\|}
\end{equation}

where $\bm{e}_\text{tex}$ and $\bm{e}_\text{vis}$ denote the text and visual embeddings produced by the VLM's text and visual encoders, respectively. At time step 100, the task instruction is changed from ``FLY past the LEFT side of the gate then FLY to LADDER base" to ``FLY past the RIGHT side of the gate then STOP over MONITOR." The resulting drone trajectory and the corresponding VLM similarity scores over time are visualized in Figure~\ref{fig:task_shift}.

As shown in Figure~\ref{fig:task_shift}, the VLM's normalized match score exhibits a significant drop shortly after the task instruction is changed at step 100, reflecting a temporary mismatch between the new instruction and the drone's current visual context. Following this drop, the score gradually recovers as the drone adjusts its behavior to align with the new instruction. This dynamic trend confirms that the VLM generates meaningful, context-aware similarity scores that reflect the semantic alignment between perception and language. Importantly, despite the abrupt mid-flight task switch, the policy is able to successfully complete the new task, demonstrating the robustness and adaptability of our framework under task-shift scenarios.

\re{
\subsection{MoE Scalability Experiment}
We further evaluated the scalability of our MoE policy by training variants with 4 and 8 experts on the multi-environment adaptation task. Following the protocol in Section~\ref{sec:exp_multi_env}, we kept each expert's architecture identical to Section~\ref{sec:method_VLM} and used top-2 expert selection. Results are shown in Table~\ref{table:moe_scale_combined}.

As the number of experts increases, we observe a mild performance gain in simulation but a clear drop in real-robot success. While additional experts improve multi-task learning capacity, they also introduce substantial computational overhead. On our platform, the 8-expert MoE ran at only ~10~Hz, causing delayed control and degraded performance. This highlights that, in real-world settings, MoE scaling is constrained by onboard compute, and further scaling would require either more powerful hardware or optimized inference pipelines.

}

\begin{table}[t]
\centering
\scriptsize
\caption{\re{MoE scalability experiment results in both simulation and real hardware. Each MoE policy network uses a different number of experts (2, 4, or 8). Simulation protocols follow Table~\ref{table:multi_env_sim}, while real-robot protocols follow Table~\ref{table:multi_env_real}.}}
\label{table:moe_scale_combined}
\begin{tabular}{c|c|c|c|c|c}
\toprule
\multirow{2}{*}{\textbf{Exp. Num.}} & \multirow{2}{*}{\textbf{Reward (Sim)}} & \multicolumn{4}{c}{\textbf{Success Rate (Left \textbar\ Mid.)}} \\
\cmidrule(lr){3-6}
 &  & Through & Right & Left & Over \\
\midrule
\multirow{2}{*}{2 (ours)} 
 & 4169.5 (Sim) & \textbf{6/6} \textbar\ \textbf{5/6} & 5/6 \textbar\ 4/6 & \textbf{6/6} \textbar\ 4/6 & 5/6 \textbar\ 4/6 \\
 & N/A (Real)    & \textbf{3/3} \textbar\ \textbf{2/3} & \textbf{3/3} \textbar\ \textbf{2/3} & \textbf{2/3} \textbar\ \textbf{3/3} & \textbf{2/3} \textbar\ \textbf{1/3} \\
\midrule
\multirow{2}{*}{4} 
 & 4280.4 (Sim) & \textbf{6/6} \textbar\ \textbf{5/6} & 5/6 \textbar\ \textbf{6/6} & 5/6 \textbar\ \textbf{5/6} & 5/6 \textbar\ 5/6 \\
 & N/A (Real)    & 1/3 \textbar\ \textbf{2/3} & 2/3 \textbar\ 1/3 & \textbf{2/3} \textbar\ 1/3 & 1/3 \textbar\ 0/3 \\
\midrule
\multirow{2}{*}{8} 
 & \textbf{4319.1} (Sim) & \textbf{6/6} \textbar\ 3/6 & \textbf{6/6} \textbar\ \textbf{6/6} & \textbf{6/6} \textbar\ \textbf{5/6} & \textbf{6/6} \textbar\ \textbf{6/6} \\
 & N/A (Real)    & 1/3 \textbar\ 0/3 & 0/3 \textbar\ 1/3 & 0/3 \textbar\ 1/3 & 0/3 \textbar\ 0/3 \\
\bottomrule
\end{tabular}
\end{table}

\section{Conclusion}
\re{In this paper, we present a fully onboard VLA framework that maps high-level natural language commands directly to low-level drone controls. The Vision-Language Model enables task-relevant decisions by grounding language in visual observations, while the MoE action head improves generalization via sparse routing. Training with DiffRL inside the 3DGS model further provides smooth gradients, improves sample efficiency, and yields policies that transfer to real hardware without additional tuning. Future directions include coupling with learned world models for long-horizon planning and greater efficiency.

\emph{Limitations:} Our policy adapts to novel combinations of sub-tasks seen individually during training (e.g., recombining known goals), but fails on instructions composed entirely of unseen sub-tasks. Thus, we do \textbf{not} claim this framework as a general solution for open-vocabulary embodied AI.
}
% \section{Acknowledgment}
% The authors would like to thank Chenghao Zhu for
% assistance with the PPO baseline code development. The authors would also like to thank Keiko Nagami and Javier Yu for fruitful discussion on FiGS and sharing with NeRF usage code. The GRaD-Nav algorithm was developed based on SHAC's open-sourced codebase \cite{xu2022accelerated}, $\beta$-VAE based CENet code was referred to \cite{DreamWaQ_repo}.

\newpage 
\bibliographystyle{IEEEtran}
\bibliography{ref}
\section{Appendix} \label{sec:appendix}
\setlength{\tabcolsep}{3pt}

\begin{table}[h]
    \centering
    \scriptsize
    \caption{High level natural language instructions table for drone flight VLA tasks in Section~\ref{sec:task_def}.}
    \label{table:task_instruc}
    \begin{tabular}{cc}
        \toprule
        \textbf{Stage \#1}                                   & \textbf{Stage \#2}                  \\
        \midrule
        \multirow{3}{*}{GO THROUGH gate}                     & \multirow{4}{*}{STOP over MONITOR}  \\
                                                             &                                     \\
                                                             &                                     \\
        \multirow{3}{*}{FLY past the RIGHT side of the gate} &                                     \\
                                                             & \multirow{4}{*}{STOP over CART}     \\
                                                             &                                     \\
        \multirow{3}{*}{FLY past the LEFT side of the gate}  &                                     \\
                                                             &                                     \\
                                                             & \multirow{4}{*}{FLY to LADDER base} \\
        \multirow{3}{*}{FLY ABOVE gate}                      &                                     \\
                                                             &                                     \\
                                                             &                                     \\
        \bottomrule
    \end{tabular}
\end{table}

% \begin{table}[h]
%     \centering
%     \scriptsize
%     \caption{Hyper-parameters table of different training methods.}
%     \label{table:hyper-param}
%     \begin{tabular}{lccc}
%         \toprule
%         \textbf{Parameters} & \textbf{Ours} & \textbf{PPO} & \textbf{w/o MoE} \\
%         \midrule
%         Number of envs             & 128   & 128   & 128  \\
%         Discount factor $\gamma$   & 0.99  & 0.99  & 0.99  \\
%         Actor learning rate        & 3e-4  & 3e-4  & 3e-4  \\
%         Critic learning rate       & 1e-4  & 1e-4  & 1e-4  \\
%         CENet learning rate        & 5e-4  & 5e-4  & 5e-4  \\
%         GAE $\lambda$              & 0.95  & 0.95  & 0.95  \\
%         Horizon length             & 32    & 32    & 32  \\
%         Critic updates             & 16    & -     & 16  \\
%         % Clipping parameter $\epsilon$  & - & 0.1   & -  \\
%         % Entropy coefficient        & -     & 1e-3  & -  \\
%         % MoE Aux. loss weight       & 0.5    & -  & -    \\
%         % MoE balance weight         & 10     & -  & -  \\
%         % MoE entropy weight         & 0.1     & -  & -  \\
%         \bottomrule
%     \end{tabular}
% \end{table}

\end{document}